\documentclass{IEEEtran4PSCC}
\ifCLASSINFOpdf
\usepackage[pdftex]{graphicx}
\else
\usepackage[dvips]{graphicx}
\fi
\usepackage[cmex10]{amsmath}
\usepackage{stmaryrd}
\usepackage{amssymb}
\usepackage{bbm}
\makeatletter
\let\old@ps@headings\ps@headings
\let\old@ps@IEEEtitlepagestyle\ps@IEEEtitlepagestyle
\def\psccfooter#1{%
	\def\ps@headings{%
		\old@ps@headings%
		\def\@oddfoot{\strut\hfill#1\hfill\strut}%
		\def\@evenfoot{\strut\hfill#1\hfill\strut}%
	}%
	\def\ps@IEEEtitlepagestyle{%
		\old@ps@IEEEtitlepagestyle%
		\def\@oddfoot{\strut\hfill#1\hfill\strut}%
		\def\@evenfoot{\strut\hfill#1\hfill\strut}%
	}%
	\ps@headings%
}
\makeatother

\psccfooter{%
	\parbox{\textwidth}{\hrulefill \\ \small{24th Power Systems Computation Conference} \hfill \begin{minipage}{0.2\textwidth}\centering \vspace*{4pt} \includegraphics[scale=0.06]{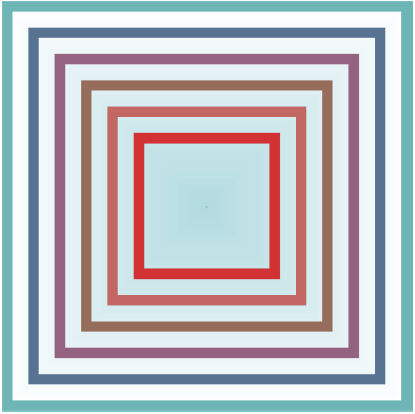}\\\small{PSCC 2024} \end{minipage} \hfill \small{Paris, France --- June 4 - 7, 2024}}%
}

\usepackage{amsmath,amssymb,amsfonts}
\usepackage{cite}
\usepackage{url}
\usepackage{subcaption} 
\usepackage[draft=False]{hyperref} 
\hypersetup{
     colorlinks=true,
     linkcolor=blue,
     filecolor=blue,
     citecolor=blue,      
     urlcolor=blue,
     }
     
\begin{document}
	%
	\title{Deep Generative Methods for Producing Forecast Trajectories in Power Systems}
	
	\author{
		\IEEEauthorblockN{Nathan Weill, Jonathan Dumas}
		\IEEEauthorblockA{RTE R\&D, LA DEFENSE, France,\{nathan.weill, jonathan.dumas\}@rte-france.com\\
			}
	}


\maketitle
	
\begin{abstract}
With the expansion of renewables in the electricity mix, power grid variability will increase, hence a need to robustify the system to guarantee its security. Therefore, Transport System Operators (TSOs) must conduct analyses to simulate the future functioning of power systems. Then, these simulations are used as inputs in decision-making processes. In this context, we investigate using deep learning models to generate energy production and load forecast trajectories. To capture the spatiotemporal correlations in these multivariate time series, we adapt \textit{autoregressive networks} and \textit{normalizing flows}, demonstrating their effectiveness against the current copula-based statistical approach. We conduct extensive experiments on the French TSO RTE wind forecast data and compare the different models with \textit{ad hoc} evaluation metrics for time series generation. 
\end{abstract}
	
\begin{IEEEkeywords}
Autoregressive models, Deep generative modeling, Energy forecasting, Multivariate time series
\end{IEEEkeywords}
	

\section{Introduction}

\begin{figure}[tb]
    \centering
    \includegraphics[scale=0.3]{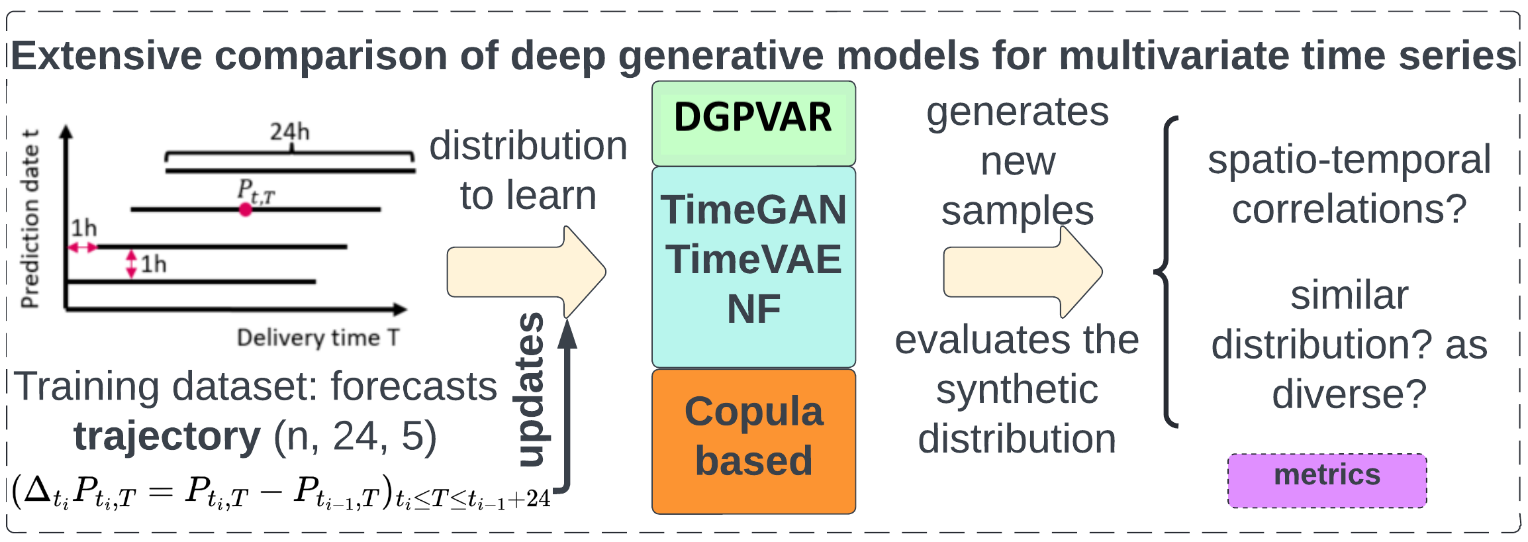}
    \caption{Framework of the paper.}
    \label{fig:schema}
\end{figure}

The problem considered in this study is called "\textit{re-forecasting}". The purpose is to generate forecast trajectories in the future for various quantities in power systems, such as renewable energy production and load, which capture the multivariate properties of operational forecasts.
We want to learn how to simulate the process of the operational forecasting method that yields intraday forecasts conditioned on past observations and weather updates.
Replicating this structure allows the generation of appropriate input data for future scenarios in system, grid, or market models, used to assess the impact of intermittent generation from renewable energy sources on decision-making processes.
Recent advances \cite{NFJonathan, DDPMJonathan} in regression-like operational energy forecasting methods using deep generative models such as \textit{variational autoencoders} (VAEs), \textit{generative adversarial networks} (GANs) or \textit{normalizing flows} (NFs) have exhibited promising results. However, the literature on simulating forecast trajectories in energy systems is relatively poor. The main difference between the two problems is that there is no conditional input on the past or meteorological conditions to generate a new multivariate time series.
This paper adopts a similar framework of the copula-based approach developed by \cite{UDE} to model the forecast updates process and to decorrelate the samples. 
Figure \ref{fig:schema} depicts the framework of the study, and the main contributions are fourfold:
\begin{enumerate}
    \item We address the crucial problem of simulating future forecast trajectories in power systems, a topic that still needs to be explored in the literature as only one statistical approach has been developed \cite{UDE}. 
    \item State-of-the-art generative models are adapted to time series and implemented, which requires a radical change in the network architectures. This study advocates for using tweaked \textit{autoregressive networks}. This class of models allows for easily considering the temporal dimension.
    \item Relevant evaluation methodologies are designed to compare the statistical approach with the time series-specific deep generative models on TSO data. 
    \item Finally, these models are tested and compared on a real-world dataset from the French TSO RTE. It comprises wind power forecast trajectories for five power substations. 
\end{enumerate}
The paper is organized as follows. Section \ref{lit} proposes a concise literature review on time series generation. Section \ref{pbst} abstractly formulates the problem and explains the modeling choices. Section \ref{meth} introduces the generative methods investigated. Section \ref{exp} describes and analyzes the numerical experiments and the evaluation metrics used. Finally, conclusions and perspectives are drawn in Section \ref{cls}.
 
\section{Literature review}\label{lit}

The literature on the "re-forecasting" problem is scarce, and to the best of our knowledge, the only study dealing with this topic is \cite{UDE}. 
It is worth mentioning that deep learning generative models have demonstrated their ability to perform regression-like operational day-ahead forecasts in recent advances \cite{NFJonathan}, \cite{DDPMJonathan} on an open access dataset (PV, wind and power consumption). However, they were not tested in a study mode forecasting context.  
Thus, we propose a broader view of time series studies presented in the literature.
Notice that we assume that the reader is familiar with basic neural network concepts, including Multi-Layer Perceptron (MLP), Recurrent Neural Networks (RNNs), and Long Short Term Memory networks (LSTMs).

\subsection{Deep Generative Modeling}

Time series generation involves the creation of synthetic data that mimics real-world time series. Since time series data is defined by its temporal interrelationships, generating such data typically involves learning the underlying patterns, correlations, and trends. In other words, given time series samples, the purpose is to learn the underlying distribution (or rather an approximation of this distribution) from which they are drawn. If a good "generative model" is learned, it could be used for downstream inference and chiefly sampling. 
Time series generation is still emerging, but image generation literature is plentiful. Generative Adversarial Networks (GANs) \cite{GAN}, Variational Autoencoders (VAEs) \cite{VAE}, and Denoising Diffusion Probabilistic Models (DDPMs) \cite{DDPM} have all been successfully used to generate high-quality synthetic images. 
There have been some attempts to replicate this success in the field of time series with TimeGAN \cite{TimeGAN}, TimeVAE \cite{TimeVAE}, and TimeGrad \cite{TimeGrad}, which adapted the previous models by adding recurrent layers, for instance. The usual shortcomings of GANs and VAEs are instability and the need to input a user-defined distribution. 
Not only are these models less mature than images, but their evaluation has also been relatively little studied. Deep generative models are notably challenging to evaluate, and the relevant metrics developed are tailored for image generation \cite{FID}. 
To the best of our knowledge, \cite{MiVo} is the only paper that compares evaluation metrics for time series generation and proposes a new method that they coin MiVo. It is a bidirectional check that evaluates the similarity between real and generated data by finding a synthetic one for each real-time series and for each synthetic time series a real one.
This metric requires a distance between time series. While it may be interesting to explore the use of alternate distances such as Dynamic Time Warping (DTW) \cite{DTW}, the paper focuses on the Euclidean distance, which is far less computationally intensive. 

Autoregressive models are another class of deep generative models that is older yet seems very promising for time series data. They factorize by the chain rule of probability the joint distribution over the n-dimensions as
\begin{align}
p(x) = \prod_i p(x_i \vert x_1, x_2, ..., x_{i-1}) = \prod_i p(x_i \vert x_{<i}).
\end{align}
Each factor of the product is named an emission probability. The model selected for this emission probability dictates the expressive power of the entire model. The ordering of the variables is natural with time series data, whereas it is often arbitrary with non-sequential data.
%
Building on Pixel models \cite{PixelCNN}, Wavenet \cite{Wavenet} has successfully managed to produce realistic sound waves (a kind of time series) by using masked dilated convolutions, also known as causal convolutions. The main problem with these models is that the inference time can be extended. 
%
The study \cite{PixelSNAIL} proposes a refined version of the previous ones, aiming at mitigating some of their limits. 
Finally, \cite{Maximilien} applies Autoregressive Implicit Quantile Networks (AIQN) \cite{aiqn} to time series. The advantage of this last model is that replacing the likelihood loss with a quantile loss stabilizes the training and allows for generating new samples conditioned on an input quantile. 

\subsection{Probabilistic Forecasting}

More extensive use of deep autoregressive models for time series can be found in the probabilistic forecasting literature, \textit{i.e.}, estimating a time series' future probability distribution given its past and potentially some exogenous variables (\textit{e.g.}, meteorological conditions). The main difference with the problem of time series generation is the presence of these conditional inputs. When learning, models often try to forecast a part of the time series given its starting other half. However, the model's architecture can still be relevant to design time series generation models. 

The main idea is an additional step between the factorization of the joint distribution and the emission probability model. It uses recurrent or self-attention layers to encode the past information $x_{<i}$ in a fixed-length vector. Then, the emission model varies. DeepAR \cite{DeepAR} uses relatively simple emission models, DGPVAR \cite{DGPVAR} uses a Gaussian copula model across the geographical dimension, and what is called RNN-NF \cite{RNN-NF} in this paper uses Normalizing Flows such as MADE or RealNVP \cite{MADE, MAF, RealNVP} to model the emission density. This last model has been extended with a dependency-graph encoding in \cite{GANF} in the context of anomaly detection in multivariate time series (a low learned density signals an anomaly).
Finally, a quantile version has also been developed \cite{MQF2}. 
Notice that the studies \cite{UDE} and \cite{RNN-NF} provide more information about Gaussian copulas. However, the paper \cite{UDE} uses the copula to model the correlations inside the time series. In contrast, the second study \cite{RNN-NF} uses it to model the spatial correlations since the autoregressive model and the recurrent layers already handle the temporal aspect of the data. 

Finally, the CRPS and its variants, the Energy Score (ES) and the Variogram Score (VS), are drawn from the probabilistic forecasting literature. They are proper scoring rules \cite{CRPS} and used in \cite{UDE} to evaluate the model. Essentially, they measure the quality of a forecast by comparing one ground truth value to a distribution of values (or empirically to a set of generated values). Their use in our specific problem is discussed later on.

\section{Problem Statement} \label{pbst}

\begin{figure}[tb]
    \centering
    \includegraphics[width=\linewidth]{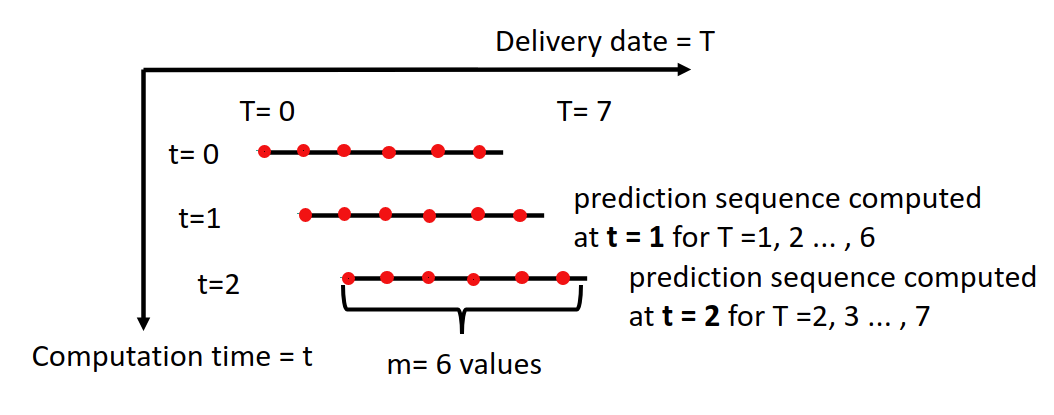}
    \caption{Illustration of three sequences of length prediction $m=6$ values. At $t=0$ a sequence is generated for forecast from $T=0$ to $T=5$. At $t=1$, a new sequence is generated from $T=1$ to $T=6$.}
    \label{fig:prediction_sequences}
\end{figure}

\begin{figure}[tb]
    \centering
    \includegraphics[width=\linewidth]{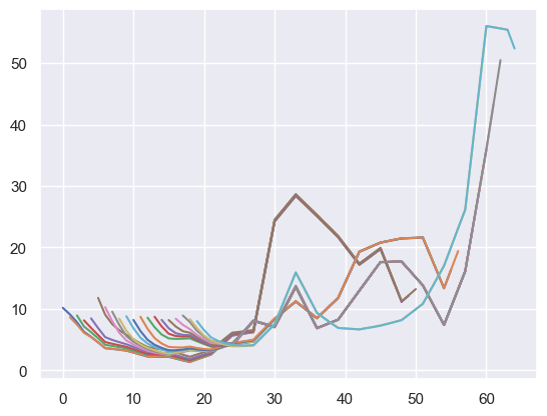}
    \caption{Example of 20 sequences of prediction of wind electricity production (y-axis MW and x-axis hours) for an electrical substation comprising several wind farms. Each sequence is composed of 46 values. The forecasting model generates a new sequence each hour, updating all the previous values and adding a new one.}
    \label{fig:prediction_train}
\end{figure}

A \textit{forecast trajectory} is a succession of $n$ forecast sequences of length $m$ moving forward in time and spanning a total period, for instance, one year. 
Every hour, the TSO forecasting model generates a new forecast sequence for the next $m$ hours, updating the latest forecast points and adding one for the new time step. 
Figure \ref{fig:prediction_sequences} illustrates three sequences of predictions, each one of length $m=6$ values, for a given area (electrical substation or wind/PV farm).
Then, these sequences of forecasts are computed for each area of interest, such as electrical substations and wind/PV farms.
Therefore, each forecast sequence comprises $m$ forecast points in a $d$-dimensional space, where this dimension represents the forecast's geographical location, \textit{e.g.}, a wind farm's electric production. 
The prediction computed at time $t$ for $T$ is $P_{t, T}$. The historical dataset comprises a forecast trajectory for $d$ electrical substations (in this study, we restrict the analysis exclusively to wind farms connected to these electrical substations). 
The $i^{th}$ sample, $P^i=(P_{t_i,T})_{t_i\leq T \leq t_i+m-1} \in \mathbb{R}^{m \times d}$, with  $0\leq i \leq n-1$, is a multivariate time series representing a $m$ hours prediction of electricity production for $d$ areas. Furthermore, each sequence's first point is considered an observation $P_{t_i,t_i} = P_{t_i} \ \forall i$.
Figure \ref{fig:prediction_train} depicts wind
electricity production forecasts in one dimension (one electric substation) with 20 prediction sequences of length $m=46$ hours.

In the vein of \cite{UDE}, instead of focusing on the different forecasts $ P_{t, T}$, we concentrate on forecast updates for a given delivery date $T$
\begin{align}
\Delta P_{t, T}=P_{t, T} - P_{t-1, T}
\end{align}
to provide a consistent probabilistic simulation of forecast trajectories.
Notice that the time length of each sequence $\Delta P_{t, T}$ is $m'=m-2$ since only the updates for the same time horizon $T$ can be computed.
Then, the following hypotheses regarding significant forecast properties of the updates are adopted to develop a model for multivariate intraday forecast update processes
\begin{subequations}
\begin{align}
    Corr(\Delta P_{t,t+k}, \Delta P_{t+i,t+k}) = 0
    \label{corrz}, \\
    Corr(\Delta P_{t,t+k}, \Delta P_{t,t+k+j}) > 0.
    \label{corrp}
\end{align}
\end{subequations}
(\ref{corrz}) means $\Delta P^i=(\Delta P_{t_i,T})_{t_i\leq T \leq t_i+m'-1} \in \mathbb{R}^{m \times d}$, for $i=0, ..., n'-1$ are assumed to be uncorrelated. Indeed, later updates of forecasts for a given delivery time should be a consequence of new information. Indeed, as a first approximation, the new information that leads to a forecast change should be unpredictable, \textit{i.e.}, the new information process is not auto-correlated. Therefore, there arguably should be no correlation between subsequent forecast updates and induction between any of the samples.
However, (\ref{corrp}) indicates that contemporaneous forecast updates (published simultaneously for different forecast horizons) should be positively correlated since the same information update induced them. We contend that this correlation constitutes a time series process.
Thus, the original problem is reframed as the learning and generation of a multivariate time series process similar to \cite{UDE}. 

\begin{figure}[tb]
	\begin{subfigure}{.5\textwidth}
		\centering
		\includegraphics[width=\linewidth]{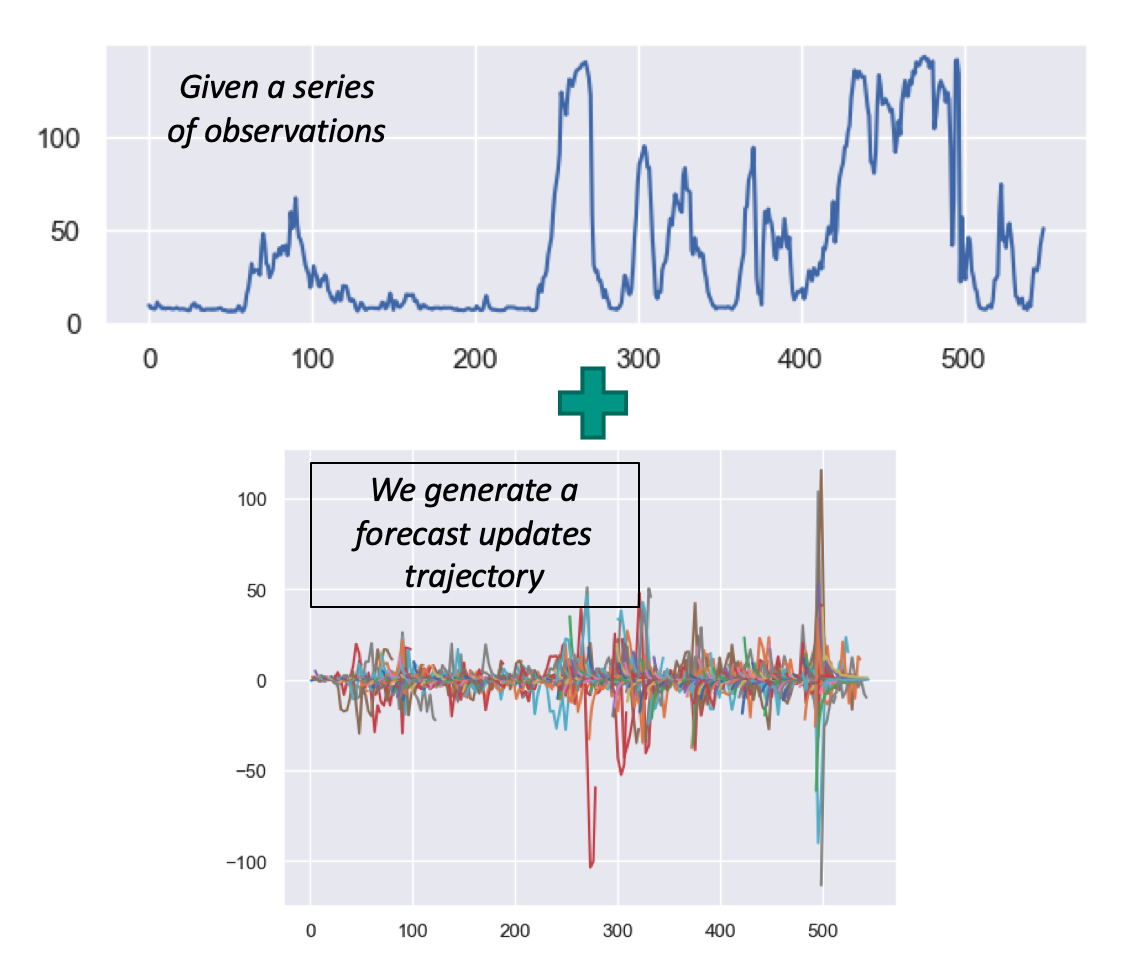}
		    \caption{From observations (pseudo-observations computed from climate models, for instance) and generated updates trajectory...}
	\end{subfigure}
 	\begin{subfigure}{.5\textwidth}
		\centering
		\includegraphics[width=\linewidth]{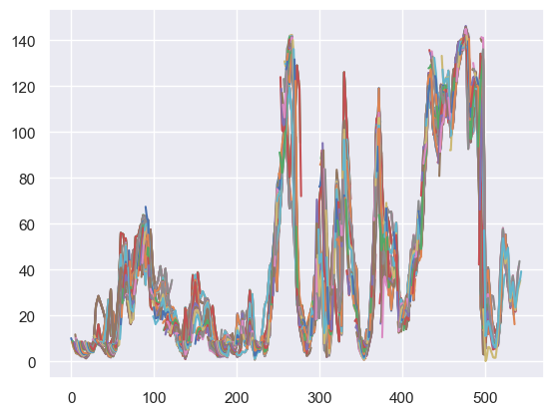}
		\caption{...we build a realistic trajectory of forecasts capturing the significant forecast properties of the actual forecasting models.}
	\end{subfigure}
	\caption{Overall process of generating a forecast's trajectory from a series of pseudo-observations.}
	\label{fig:process}
\end{figure}

From generated updates $\Delta P_{t_i, T}$ computed by a model, one can build a new forecast trajectory of length $m$ $\{P_{t, T}\}_{T-m\leq t<T}$ by cumulatively subtracting updates $\Delta P_{t_i, T}$ from a set of pseudo-observations $\Tilde{P}_T$ provided by another model, where each $P_{t, T}$ is obtained with
\begin{align}
P_{t, T} = \Tilde{P}_T - \sum_{t<i<T} \Delta P_{t_i, T}.
\end{align}
Recall that these forecast trajectories are generated in case studies where weather forecasts are unavailable, for instance, for prospective studies in 2030 or 2050. Thus, climate trajectories are computed by specific models that provide pseudo-observations $\Tilde{P}_T$ for wind and PV farms. Then, by applying $\Delta P_{t_i, T}$, it is possible to simulate realistic forecasts (re-forecasts) of wind and PV generation.
Figure \ref{fig:process} depicts this overall process of rebuilding a forecast's trajectory from pseudo-observations.

\section{Proposed Methods}\label{meth}

This study proposes challenging the copula-based approach \cite{UDE} with three deep learning-based generative models: a normalizing flow and two autoregressive models. 

\subsection{Copula-based model}

For more details about the model developed by UDE, please report to their paper \cite{UDE}. We only define what a copula is to keep this paper self-contained.
A copula function $C : [0, 1]^N \rightarrow [0, 1]$ is the cumulative distribution function (CDF) of a joint distribution of a collection of real random variables $U_1, ..., U_N$ with uniform marginal distribution, \textit{i.e.}, $C(u_1, ..., u_N ) = P(U_1 \leq u_1, ..., U_N \leq u_N)$. The critical result is Sklar’s theorem, which states that any joint cumulative distribution $F$ admits a representation in terms of its univariate marginals $F_i$ and a copula function $C$
\begin{align}
   F(z_1, . . . , z_N ) = C\big(F_1(z_1), . . . , F_N(z_N)\big).
\end{align}
When the marginals are continuous, it also states that the copula $C$ is unique and is given by the joint distribution of the probability integral transforms of the original variables.

Notice that UDE developed a spatial extension of the copula-based approach \cite{UDE} capable of simultaneously considering several areas such as electrical substations and wind farms. This extension is compared to the three deep-learning generative models proposed in this paper.

\subsection{Normalizing Flows}

The first deep-learning generative-based model considered is a normalizing flow \cite{MAF}. 
It learns an invertible transformation between the data and some Gaussian noise. Applying the inverse transformation to newly sampled Gaussian noise generates new synthetic data. 
This study implemented a straightforward normalizing flow, a univariate model, as a baseline. To use it to learn the multivariate time series, the different spatial dimensions were concatenated along the same axis for each sample. They were normalized individually and thus kept their amplitude difference on average, yielding better results than a uniform normalization. 
The normalizing flow developed by \cite{GANF} (either the MAF, MADE, or the RealNVP flow) is directly available on Github\footnote{\url{https://github.com/EnyanDai/GANF/blob/main/models/NF.py}}. 

\subsection{Autoregressive models}

The second deep-learning generative-based model is the Deep Gaussian Process Vector Auto-Regressive (DGPVAR) model \cite{DGPVAR}. Thereafter, the main equations explaining the model are presented. 
For a sample $z = (z_1, ..., z_m)$ of the dataset where each timestep $z_i \in \mathbb{R}^d$ where $d$ is the spatial dimension
\begin{align}
    p(z) = \prod_{i=1}^m p(z_i \vert z_1, ..., z_{i-1}) = \prod_i p(z_i \vert z_{<i}) = p(z_i \vert h_i),
\end{align}
where the past encoding is achieved with a recurrent layer (an LSTM in practice) 
\begin{align}
h_i = RNN(h_{i-1}, z_{i-1}) \in \mathbb{R}^D, \quad \forall i.
\end{align}
Then, the emission probability is modeled with a Gaussian Copula
\begin{align}
p(\mathbf{z}_i | \mathbf{h}_i)=\mathcal{N}\big([f_1(z_{1, i}), ..., f_d(z_{d, i})]^T | \boldsymbol{\mu}(\mathbf{h}_i), \Sigma(\mathbf{h}_i)\big),
\end{align}
with $(f_j)_{j=1,...,d}$ invertible mappings of the form $\phi^{-1} \circ \hat{F_j}$, where $\phi$ denotes the cumulative distribution function of the standard normal distribution, and $\hat{F_j}$ denotes an estimate of the marginal distribution of the i-th time series $z_{1,i}, . . . , z_{d,i}$. The estimate used was the empirical cumulative distribution function defined as
\begin{align}
\hat{F_j}(v) = \frac{1}{m} \sum_{i=1}^m \mathbbm{1}_{z_{j,i}\leq v}.
\end{align}
These functions transform the data for the i-th time series so that it follows a standard normal distribution marginally.
Finally, functions $\mu(\cdot)$ and $\Sigma(\cdot)$ map the state $h_i$ to the mean and covariance of a Gaussian distribution over the transformed observations.
The covariance matrix is modeled using a low-rank parametrization $\Sigma = D + VV^T$ where $D \in \mathbb{R}^{d \times d}$ is diagonal and $V \in \mathbb{R}^{d\times r}$. This is useful when the problem is high-dimensional. 
Discussion about the low-rank parameter $r$ can be found in \cite{DGPVAR}. Finally, the expressions of the mappings are
\begin{subequations}
\begin{align}
    \mu_j(h_{j, i}) & = w_{\mu}^T h_{j,i}, \\
    d_j(h_{j, i}) & = s(w_{d}^T h_{j,i}), \\
    v_j(h_{j, i}) & = W_v h_{j,i},
\end{align}
\end{subequations}
with $s(x) = \log (1 + e^x)$ the soft-plus function, and $w_{\mu} \in \mathbb{R}^d, w_{d} \in \mathbb{R}^d, W_{v} \in \mathbb{R}^{r\times d}$ are shared parameters across the spatial dimensions. 

The copula-induced expressions and the training phase have been adapted to the generative setting instead of the forecasting setting considered in \cite{DGPVAR}.
The Gaussian Copula parameters are learned in addition to those of the recurrent network. Then, the corresponding optimal parameters are obtained by maximizing the log-likelihood using stochastic gradient descent-based optimization
\begin{align}
-\log[p(z_1, ..., z_m)] = - \sum_{i=1}^m \log[p(z_i|h_i)].
\end{align}
For inference, to generate new samples from the learned model, we can sequentially sample $z_i|h_i$ from the conditional distribution and update the recurrent state encoding the past information $h_i$. 

The last model experimented in-depth with is the RNN-NF \cite{RNN-NF} model, which replaces the Gaussian copula emission probability part by stacked conditional normalizing flows such as RealNVP or MAF. The density estimation can be written as
\begin{align}
\log p(z_i|h_i) = \log q(f(z_i, h_i)) + \log |\det \nabla_{z_i} f(z_i, h_i)|,
\end{align}
where $q$ is chosen as the standard normal density and $f : \mathbb{R}^d \times \mathbb{R}^D \rightarrow \mathbb{R}^d$ is a conditioned normalizing flow model such as MAF.

\section{Numerical Experiments} \label{exp}

The numerical experiments are conducted on historical operational data from the French TSO RTE. The three models using deep learning generative models (NF, DGPVAR, and RNN-NF) are compared to the copula-based approach of \cite{UDE} using quality metrics.

\subsection{Case study}

The numerical experiments are conducted on a case study composed of five electrical substations of the French TSO RTE. Each substation encompasses several windmills or wind farms from a similar geographical location.
The wind power forecast trajectories computed in 2021 for these five substations were considered. 
The dataset spanned 2021 at an hourly timestep, with every prediction sequence composed of 46 values, using the earlier notations $n=8712, m=46, d=5$. 
The observation is considered the prediction computed at $T$ for $T$. 

\begin{figure}[tb]
    \centering
    \includegraphics[width=\linewidth]{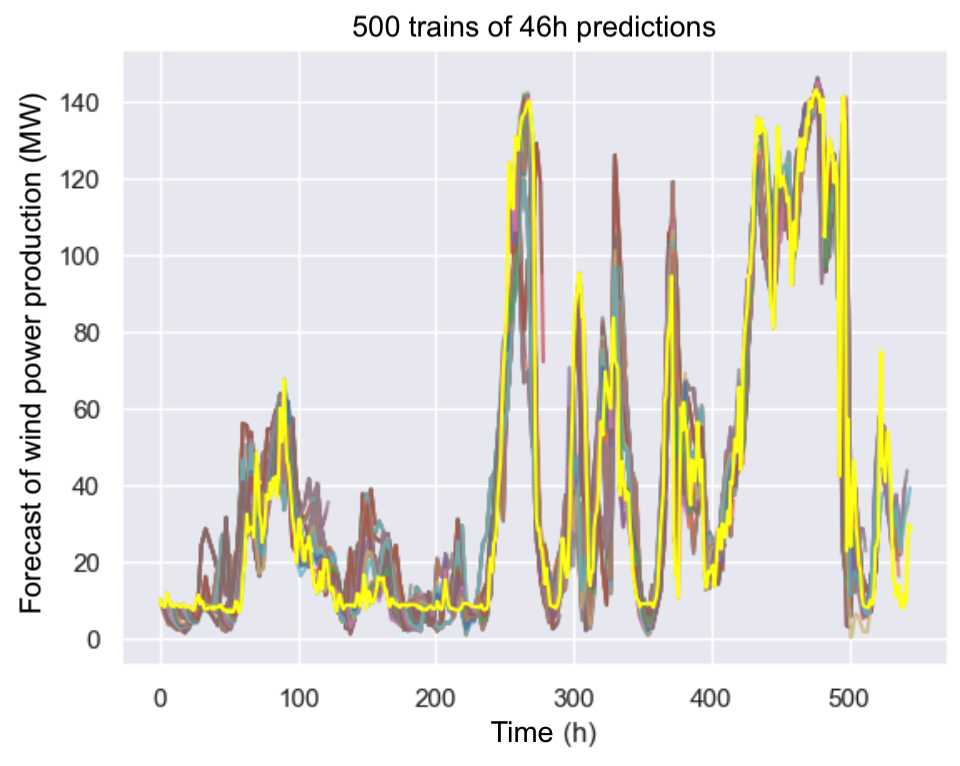}
    \caption{500 prediction sequences of 46 hours each and the observations (yellow).}
    \label{fig:data}
\end{figure}

As per the computation time, with a machine running on an NVIDIA GPU RTX A1000, the training for 11 months took approximately 20 minutes, and the generation of 10 scenarios and rebuilding of the associated trajectories took 15 minutes.
Figure \ref{fig:data} depicts 500 prediction sequences of 46 hours each.

\subsection{Evaluation metrics from two angles: MiVo and ES/VS}

\begin{figure}[tb]
    \centering
    \includegraphics[width=\linewidth]{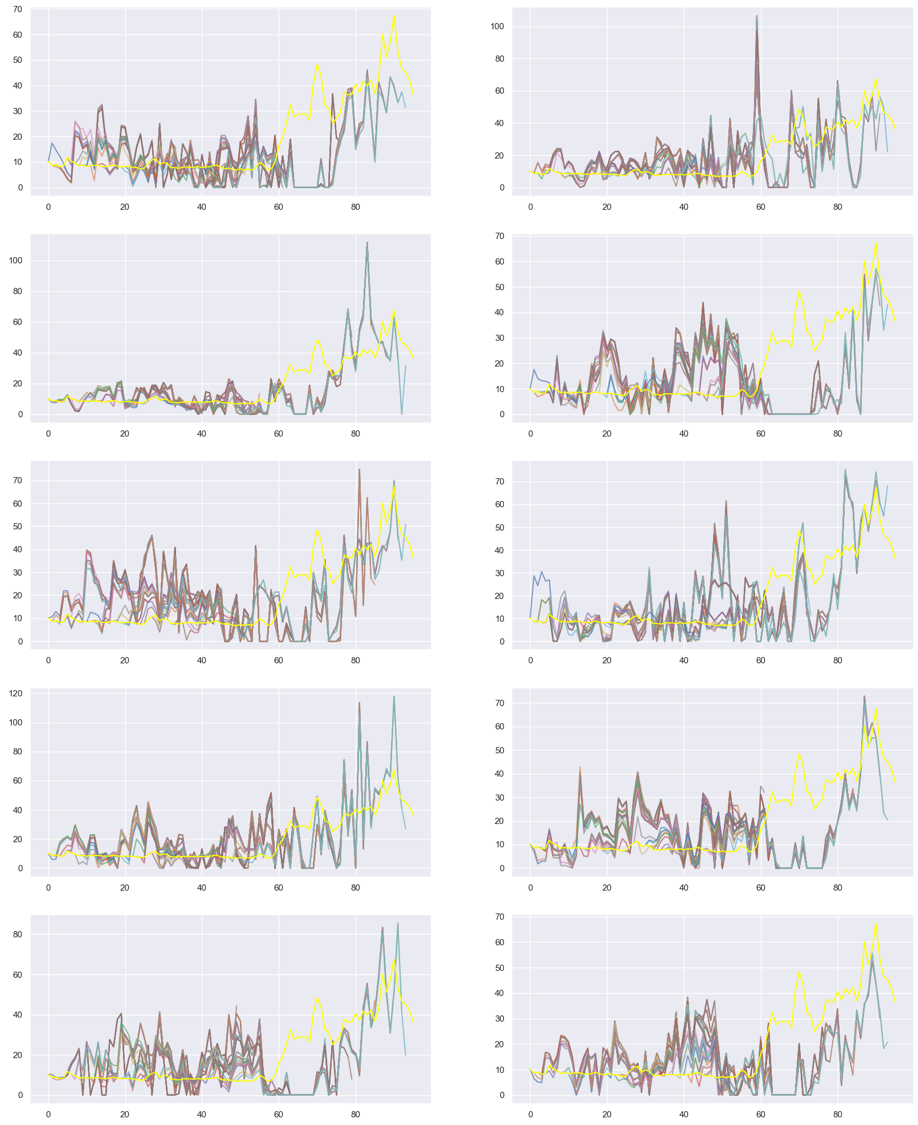}
    \caption{10 scenarios of prediction sequences generated by the simple NF approach (y-axis MW and x-axis hours) for a given electrical substation of the dataset comprising several wind farms. The observations are in yellow.}
    \label{fig:10scen}
\end{figure}

\begin{figure}[tb]
    \centering
    \includegraphics[width=\linewidth]{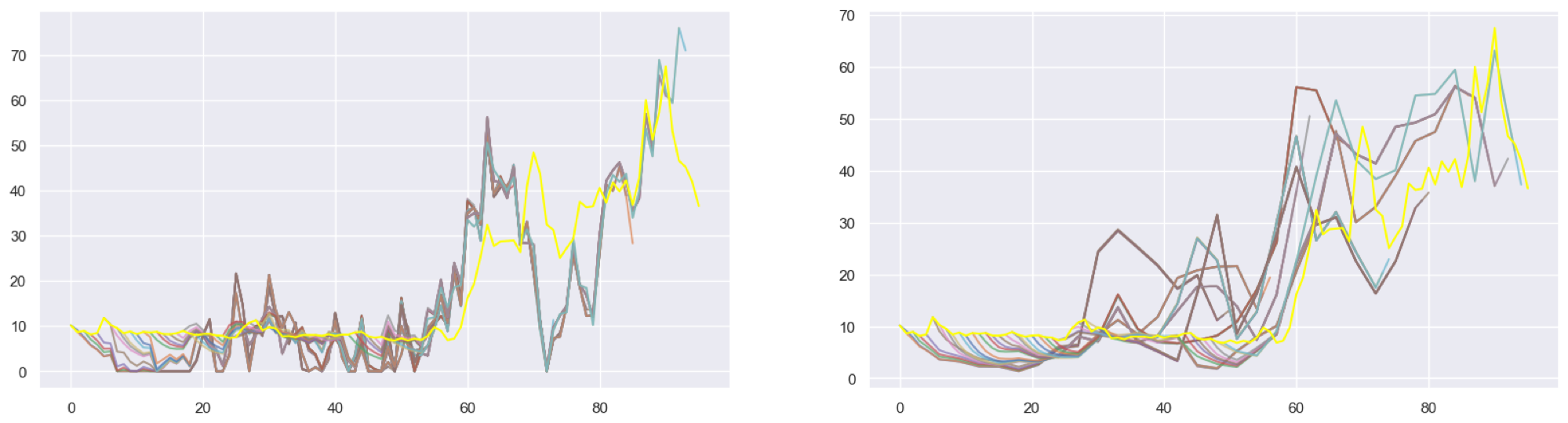}
    \caption{Comparison of generated (left) and historical (right) sequences of prediction (y-axis MW and x-axis hours). The observations are in yellow.}
    \label{fig:comp}
\end{figure}

The updates generated can be used directly to compute the MiVo metric.
Following \cite{MiVo}, a good model can generate realistic data if each generated time series achieves a low minimal distance to at least one real-time series. In addition, it should generate diverse samples if, for each real-time series, it is possible to find a generated partner.
By noting, $D=(d_{ij})_{i,j}$, the distance matrix between the two sets of real and generated update time series, the MiVo is $\mu(D_1) + \sigma^2(D_2)$, where $\mu$ is the mean and $\sigma^2$ is the variance, and 
\begin{subequations}
\begin{align}
D_1 = & \big[\min (d_{11}, \ldots, d_{1n}), \ldots, \min (d_{m1}, \ldots, d_{mn})\big], \\
D_2 = & \big[\min (d_{11}, \ldots, d_{m1}), \ldots, \min (d_{1n}, \ldots, d_{mn})\big].
\end{align}
\end{subequations}
In contrast to the MiVo metric, several scenario trajectories must be rebuilt first to compute the ES and VS scores \cite{CRPS}. 
Figure \ref{fig:10scen} depicts ten trajectories generated by the NF model, and Figure \ref{fig:comp} emphasizes the difference between the natural and synthetic data.
Then, for each sample $t$ composed of $m$ values and area $r$, the energy score is computed as follows by using $S$ scenarios
\begin{align}
\widehat{E S}_{r, t} = & \frac{1}{S} \sum_{s=1}^S \sqrt{\sum_{k=1}^m \left(\hat{p}_{r, t, t+k}^s-p_{r, t+k}^0\right)^2} \nonumber \\
& -\frac{1}{2(S-1)} \sum_{s=1}^S  \sqrt{\sum_{k=1}^m \left(\hat{p}_{r, t, t+k}^s-\hat{p}_{r, t, t+k}^{s+1}\right)^2},
\end{align}
with $p_{r, t}^0 = p_{r,t,t}$. 
Then, the final aggregated energy score is $\widehat{E S} = \frac{1}{R N_t} \sum_{r, t} \widehat{E S}_{r, t}$, with $R$ the number of areas considered, and $N_t$ the sample size.
Similarly, the Variogram score is computed as follows with $\gamma=1$
\begin{align}
V S_T = & \sum_{r_1, r_2} w_{r_1, r_2}\bigg(|p_{r_1, T}^0-p_{r_2, T}^0|^\gamma \nonumber \\
& -\frac{1}{K S} \sum_{k=1}^m \sum_{s=1}^S|\hat{p}_{r_1, T-k, T}^s - \hat{p}_{r_2, T-k, T}^s|^\gamma\bigg)^2.
\end{align}

\subsection{Results}

Table \ref{tab:results} provides the scores obtained for the models. 
The MiVo metric is computed on the updates, and the ES and VS are computed on 100 scenarios averaged monthly. 
\begin{table}[tb]
\renewcommand{\arraystretch}{1.25}
\begin{center}
\begin{tabular}{lrrrr}
\hline \hline
 Score & Copula-based approach & NF & DGPVAR & RNN-NF\\
 \hline
 ES   & \textbf{4.284} & 4.321 & 4.295 & 4.302\\
 \hline
 VS &   \textbf{12.310} &  12.983 & 12.534 & 12.768\\
 \hline
 MiVo &809 &  110 & \textbf{48} & 73 \\
\hline \hline
\end{tabular}
\caption{Quality evaluation of the models. The lower the score, the better. The Copula-based approach is the spatial extension of \cite{UDE} and is developed by UDE, which implemented it with a t-copula.}
\label{tab:results}
\end{center}
\end{table}
The best-performing model for the MiVo score is the DGPVAR, while the extension of the copula-based model from \cite{UDE} implemented with a t-copula achieved the best values for the energy and variogram scores. 
DGPVAR is the most successful among the three deep-learning generative models developed in this study. It achieves a trade-off between complexity and expressiveness. The number of layers inside the model has been decreased to obtain the convergence of the validation loss. In other words, a more complex model such as RNN-NF had too many parameters to learn correctly, even in its most basic version. Conversely, a simple model like NF could also not capture the essence of the distribution.

The copula-based model's superior performance on the ES and VS scores was unsurprising. Indeed, these metrics are designed to evaluate the quality of a forecasting model. The historical forecast trajectory is not involved in the computation of the scores, only the historical observations. 
In addition, some layers of complexity were added when rebuilding the trajectories, which could explain its better performance. 
Nevertheless, at its core, these metrics are flawed for the problem of re-forecasting, which aims at reproducing the forecasting process, including its statistical forecasting errors. The MiVo score tries to measure that by comparing the synthetic and historical forecasting trajectories with no regard for the actual realizations used to rebuild the trajectories. 
However, the ES and VS scores have the merit of evaluating the final rebuilt trajectories, whereas the MiVo makes sense only when computed on the updates directly. 

\subsection{Limitations of the current approach}

The proposed approach comprises several limitations.
First, it does not allow to bound the generated forecasts $P_{t, T}$ directly between $0$ and a maximum power. 
This drawback is overcome by clipping the values when constructing the trajectory from the generated updates and the observations. However, further work is required to address this issue better.
Second, the model proposed yields time series that are less smooth than the original ones, owing to the very nature of the reconstruction. Nevertheless, a carefully designed post-processing step should solve this. 
Finally, the main problem encountered was needing more data, as often with machine learning models with many parameters. It was problematic as only a sixth of the data was stochastic and contained information. Indeed, the operational forecast model uses updated weather forecasts every six hours. In between, the other five updates are only Kalman-like informational updates, affecting the forecasts in a decreasingly exponential manner \textemdash the new observation will not change the long-term forecasts unless the model was completely off, which is not the case. We adopted a parametric exponential model for this update to better account for this phenomenon. 

\section{Conclusion} \label{cls}

The problem of re-forecasting is still an open problem. The literature is still scarce, and more needs to be done to understand its ins and outs better. 
This study adopted the framework of \cite{UDE} with the generation of updates and the cumulative rebuilding of trajectories.
It introduced deep learning models to capture the spatiotemporal correlations in the multivariate time series of wind power productions by adapting autoregressive networks and normalizing flows.
It demonstrated their effectiveness against the spatial extension of the copula-based approach developed by \cite{UDE} with extensive experiments on the French TSO RTE wind forecast data.

However, this framework of generation of updates is arguable and raises additional issues. 
First, the generated values are not necessarily bounded as physical constraints dictate. 
Second, the user must have a historical trajectory of forecasts to train the model and pseudo-observations in the future to project the generated updates. It means that another model has to be developed to consider a macro perspective on the future behavior of the electrical grid and the meteorological conditions, which would radically change in thirty years. 
Third, the method of rebuilding trajectories from updates yields less smooth trajectories. 
Fourth, the assumption of non-correlation between the forecast updates could be challenged to achieve better results with generative models. 
Last but not least, this study was hindered by the lack of data. The same models with more layers and thus parameters, or even somewhat more complex models that extend this framework, such as GANF \cite{GANF}, should bring better results were more data available. In the meantime, one could look for data augmentation strategies and techniques to reduce overfitting. Dropout was used in the different networks, but spectral regularization is an interesting point to be explored. We leave that for future studies.

\section*{Acknowledgment}

The authors would like to acknowledge the work of M.Sc. Aiko Schinke-Nendza, M.Sc. Yannik Pflugfelder and Prof. Dr. Christoph Weber at UDE which developed the spatial extension of the copula-based approach \cite{UDE}. The reference of the working paper presenting this extension will be included when published. 

\bibliographystyle{IEEEtran}
\bibliography{biblio}

\end{document}